\documentclass[a4paper, 10 pt, conference]{ieeeconf}      

\IEEEoverridecommandlockouts                              

\usepackage{amsmath}
\usepackage{graphicx}
\usepackage{subcaption}
\usepackage[dvipsnames]{xcolor}
\usepackage{tabularx}
\usepackage{accents}
\usepackage{harpoon}
\usepackage{gensymb}
\usepackage{cite}
\usepackage{caption}
\DeclareCaptionFont{10pt}{\fontsize{10pt}{10pt}\selectfont}
\newcommand{\robot}{HISSbot}

\newcommand{\sidewind}{sidewinding control}

\usepackage[capitalize,nameinlink]{cleveref}
\crefformat{equation}{(#2#1#3)}
\Crefname{table}{TABLE}{TABLES}
\crefformat{section}{\S#2#1#3}
\crefformat{subsection}{\S#2#1#3}
\crefformat{subsubsection}{\S#2#1#3}
\crefrangeformat{section}{\S\S#3#1#4 to~#5#2#6}
\crefmultiformat{section}{\S\S#2#1#3}{ and~#2#1#3}{, #2#1#3}{ and~#2#1#3}
\newcolumntype{Y}{>{\centering\arraybackslash}X}

\title{\LARGE \bf
HISSbot: Sidewinding with a Soft Snake Robot
}

\author{Farhan Rozaidi$^{1}$, Emma Waters$^{2}$, Olivia Dawes$^{3}$, Jennifer Yang$^{4}$, Joseph R. Davidson$^{1}$, and Ross L. Hatton$^{1}$
\thanks{$^{1}$Farhan Rozaidi, Joseph R. Davidson, and Ross L. Hatton are with the Collaborative Robotics and Intelligent Systems (CoRIS) Institute,
        Oregon State University, Corvallis OR, USA.
        {\tt\small \{nikahman,joseph.davidson,
        ross.hatton\}@oregonstate.edu}}%
\thanks{$^{2}$Emma Waters is with Bard College at Simon's Rock,
        Great Barrington, MA, USA.
        {\tt\small  ewaters18@simons-rock.edu}}%
\thanks{$^{3}$Olivia Dawes is with Olin College of Engineering,
        Needham, MA, USA.
        {\tt\small odawes@olin.edu}}%
\thanks{$^{4}$Jennifer Yang is with Carnegie Mellon University,
        Pittsburgh, PA, USA.
        {\tt\small jennifey@andrew.cmu.edu}}%
}

\begin{document}

\maketitle
\thispagestyle{empty}
\pagestyle{empty}

\begin{abstract}

Snake robots are characterized by their ability to navigate through small spaces and loose terrain by utilizing efficient cyclic forms of locomotion. Soft snake robots are a subset of these robots which utilize soft, compliant actuators to produce movement. Prior work on soft snake robots has primarily focused on planar gaits, such as undulation. More efficient spatial gaits, such as sidewinding, are unexplored gaits for soft snake robots. We propose a novel means of constructing a soft snake robot capable of sidewinding, and introduce the Helical Inflating Soft Snake Robot (\robot{}). We validate this actuation through the physical \robot{}, and demonstrate its ability to sidewind across various surfaces. Our tests show robustness in locomotion through low-friction and granular media.
\\
\end{abstract}

\begin{keywords}

Soft robot materials and design, biologically-inspired robots, soft sensors and actuators
\end{keywords}

\section{INTRODUCTION}\label{sec:introduction}

Using their elongated and flexible bodies, snakes are capable of traversing through different environments. Snake robots capable of mimicking the behavior of their biological counterparts have been proposed as systems that could traverse rough environments. Most research into snake robotics has focused on serial chains of servomotors encased in rigid enclosures \cite{marvi_sidewinding_2014,wright_design_2012,kamegawa_realization_2009}, because of the ability to directly command an arbitrary shape. \textit{Soft snake robots} are composed of relatively inexpensive materials and can better mimic the smooth bodies of snakes. In addition, the compliance of soft robots provides resilience when exploring uneven and narrow passageways.

Prior studies on soft snake robots have primarily focused on planar bending, limiting the possible gaits to undulating motions \cite{onal_autonomous_2013, soft-snakes, rafsanjani_kirigami_2018}. Planar gaits are defined by body changes through a space of 2D shapes, which include gaits such as lateral undulation and inchworming. Planar gaits do not provide efficient locomotion in many real-world applications, such as on low-friction or granular terrain. Spatial gaits operate within the 3D space, and are of particular interest for their applications in traversing uneven terrain or typically inaccessible spaces \cite{hawkes_soft_2017}. 

One of the most efficient gaits utilized by snakes is sidewinding: the rolling static ground contacts transfer mass from one contact to another and minimize slip on the substrate \cite{sidewinding-on-slopes}. To the best of our knowledge, sidewinding requires 3D bending with a non-trivial phase relationship between the bending at different points along the body, which has not previously been achieved in soft snake robots.

\begin{figure}
    \centering
    \includegraphics[width=\columnwidth]{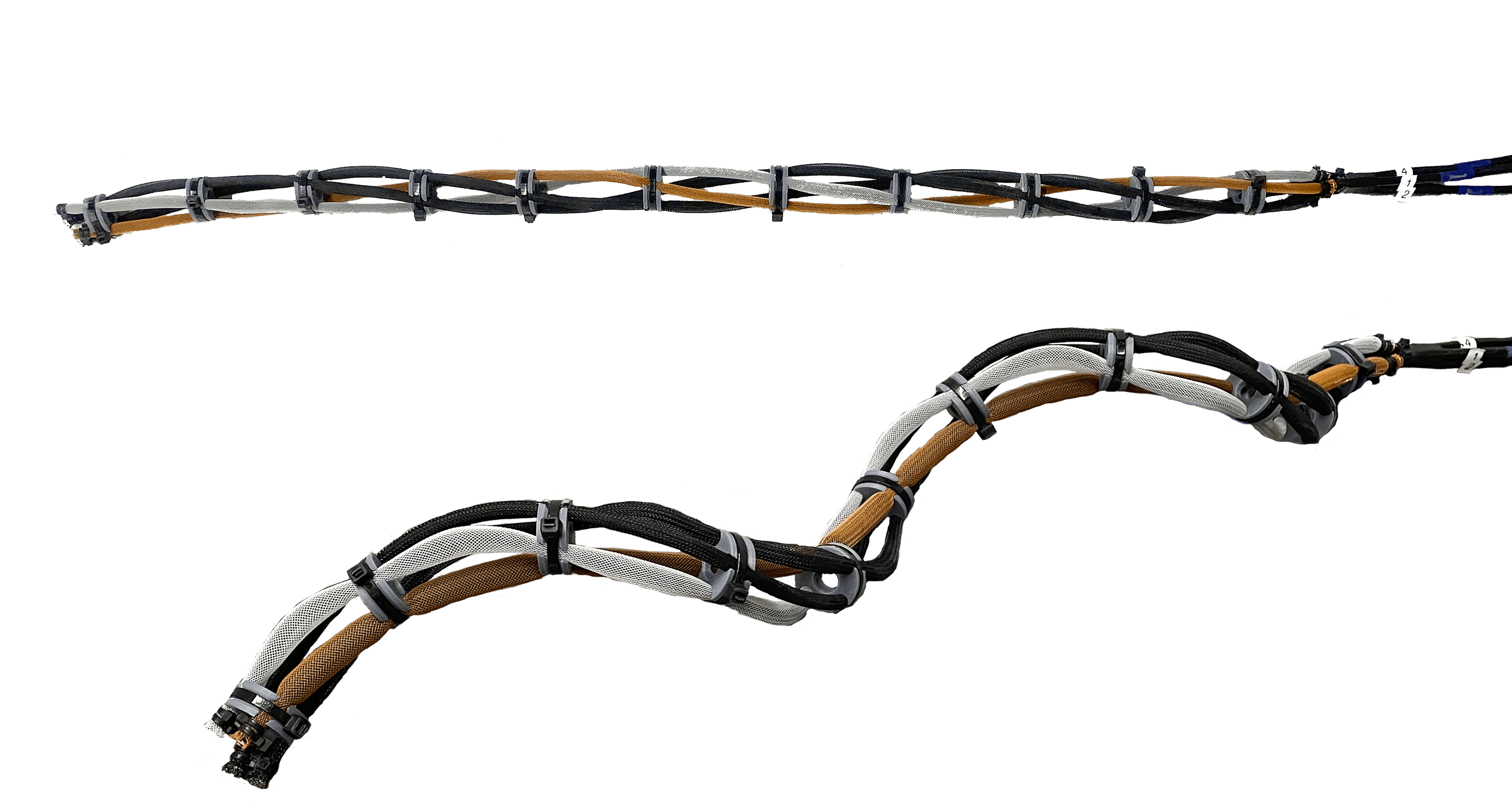}
    \caption{The \robot{} uninflated (top), and pressurized to form a helical shape for sidewinding (bottom). Cyclical contraction of the four McKibben actuators produces locomotion via a twisting-helix motion.}
    \label{fig:hissbot}
\end{figure}

In this paper, we present a Helical Inflating Soft Snake Robot (\robot{}) capable of sidewinding, as seen in \cref{fig:hissbot}. We construct a physical \robot{} with four parallel, helical, McKibben actuators to generate the helical shapes, while using a minimal number of actuators. We develop an actuation method for sidewinding, and subject the \robot{} to varying terrain, demonstrating its effectiveness in traversing through slippery or loose ground.

\section{RELATED WORK}

\subsection{Sidewinding}

\begin{figure}
    \centering
    \includegraphics[width=0.97\columnwidth]{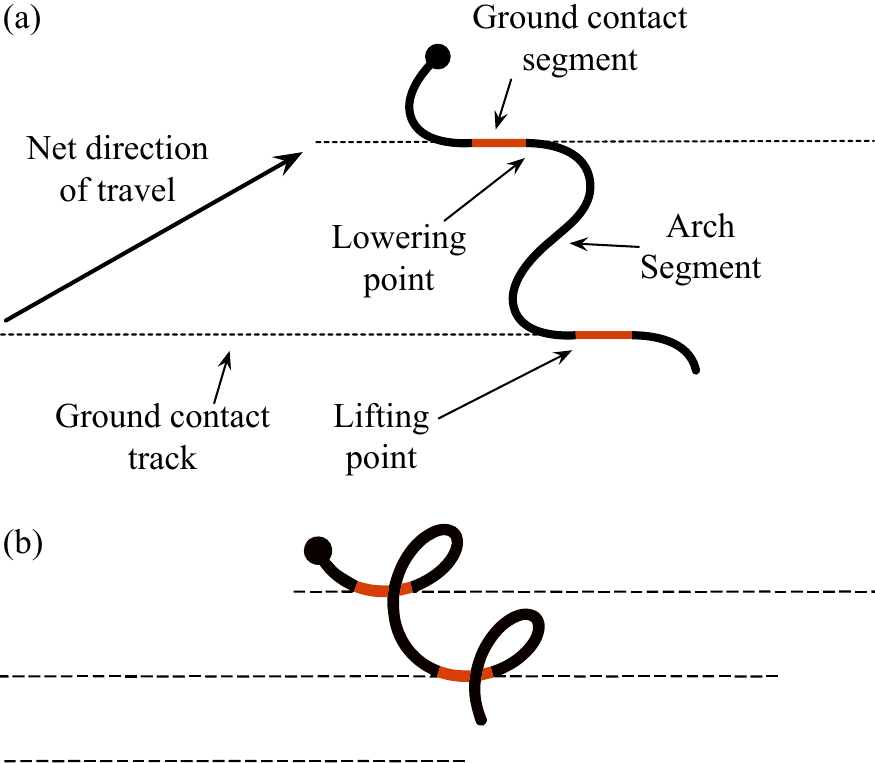}
    \caption{(a) A representation of the mechanics of a sidewinding motion: Each ground contact segment remains static while the body performs a local rolling movement by lifting and lowering its body. Through this cyclical motion, the body achieves net locomotion in a particular direction across a surface. (b) The same representation in an isometric view to emphasize the helical body \cite{sidewinding-on-slopes}.}
    \label{fig:sidewinding}
\end{figure}

Sidewinding is a highly efficient gait, initially described in the scientific literature by Mosauer \cite{mosauer_note_1930}. During sidewinding, a snake forms its body into a helical structure, and then raises and lowers portions of the helix (like the tread of a tracked vehicle) to move forward while maintaining static contact with the ground. As the body sidewinds, it lifts from one ground contact and lowers onto the next contact. These contact points trace out parallel tracks, and the body moves linearly, as illustrated in \cref{fig:sidewinding} \cite{sidewinding-on-slopes}. The static interaction of the ground contacts allow for a type of rolling-contact movement which makes sidewinding desirable for uneven terrain. 

Chirikjian and Burdick \cite{chirikjian_modal_1994} further analyzed the mechanics of sidewinding through the backbone modal approach. Tesch et al. \cite{tesch_parameterized_2009} parameterized various gaits, including sidewinding, for rigid snake robots. Hatton and Choset \cite{sidewinding-on-slopes} developed a model of sidewinding as a helical “virtual tread” and explored how flattening the geometry of the tread could provide the sidewinder with stability on slopes. Gong et al. \cite{conical-sidewinding} explored sidewinding displacement through conical helical bodies, and showed a net curved direction of travel. Marvi et al. \cite{marvi_sidewinding_2014} demonstrated sidewinding in sloped granular media through increased body contact with the surface. Sidewinding has been studied extensively in rigid snake robots, as the servomotor modules provide the actuator resolution and orientation required to create helical structures \cite{wright_design_2012,kamegawa_realization_2009,rollinson_design_2014}. To the best of our knowledge, sidewinding has yet to be achieved in soft snake robots due to the limited spatial resolution of soft actuators compared to conventional servomotors.

\subsection{Soft Snake Robots}\label{sec:softrobots}

Soft snake robots typically consist of parallel straight inflatable bladders, which either extend or contract under pressure, depending upon the type of actuator. The change of length between bladders causes bending of the body. Cyclical actuation can result in different gaits, such as lateral undulation \cite{soft-snakes} and rectilinear locomotion \cite{rafsanjani_kirigami_2018}. Most soft snake robots utilize planar bending, which makes them unable to perform gaits requiring spatial bending, such as sidewinding. Recent work by Arachchige et al. \cite{arachchige} has shown some out-of-plane bending to perform a rolling gait. However, the helical structure required to perform sidewinding has to the best of our knowledge not previously appeared in the literature, and is an unexplored gait of interest for soft snake robots.

\subsection{McKibben Actuators}

McKibben actuators are a form of pneumatic muscles introduced in the scientific literature by Joseph L. Mckibben \cite{hawkes_hard_2021,mckibben_more_1958}. These actuators consist of flexible tubing sheathed in helical interwoven sleeving. As the flexible tubing is inflated, it expands radially within the sleeving, and the interface with the sleeving results in lengthwise contraction. A robot assembled of parallel McKibben actuators can exhibit planar or spatial bending, depending on its construction. Due to the McKibben actuator's compliance and bending abilities, soft snake robots and soft robot arms have been developed with McKibben actuators \cite{arachchige,mckibben-arm,hao_design_2018}.


\section{GEOMETRY OF HELICAL SIDEWINDING}

\begin{figure}
    \centering
    \includegraphics[width=\columnwidth]{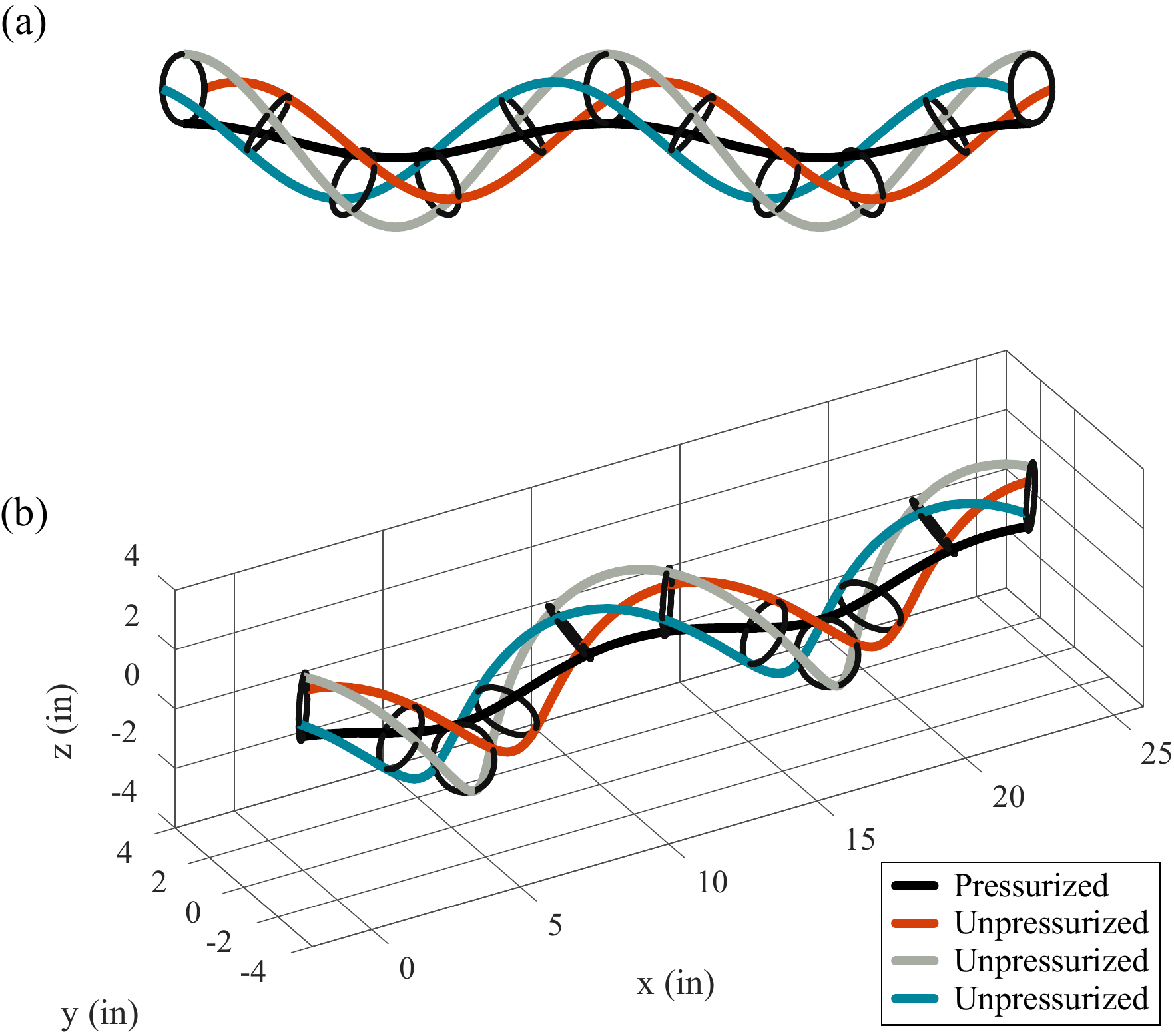}
    \caption{(a) A side view of the \robot{} with the black actuator pressurized. The model uses geometric constraints to capture the macrohelix shape during actuation. (b) An isometric view of the \robot{}.}
    \label{fig:constraintModel}
\end{figure}

\begin{figure*}
    \centering
    \begin{subfigure}[t]{0.67\columnwidth}
    \centering
        \includegraphics[width=0.95\columnwidth]{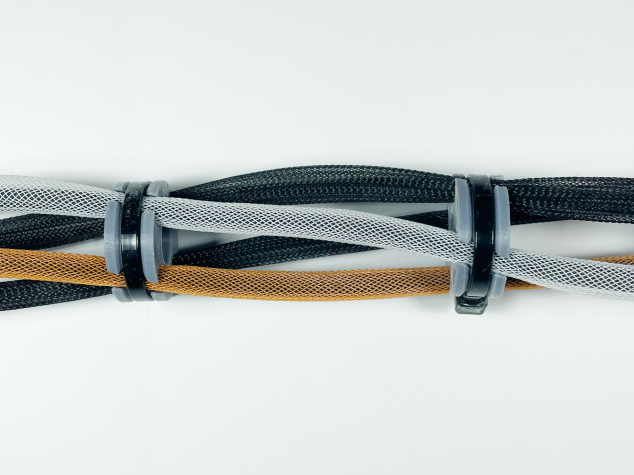}
        \caption{}
        \label{fig:snakes}
    \end{subfigure}
    \begin{subfigure}[t]{0.67\columnwidth}
    \centering
        \includegraphics[width=0.95\columnwidth]{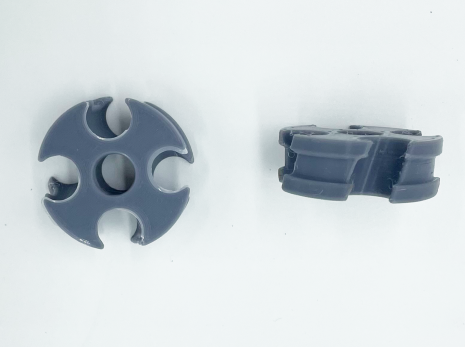}
        \caption{}
        \label{fig:braces}
    \end{subfigure}
    \begin{subfigure}[t]{0.67\columnwidth}
    \centering
        \includegraphics[width=0.95\columnwidth]{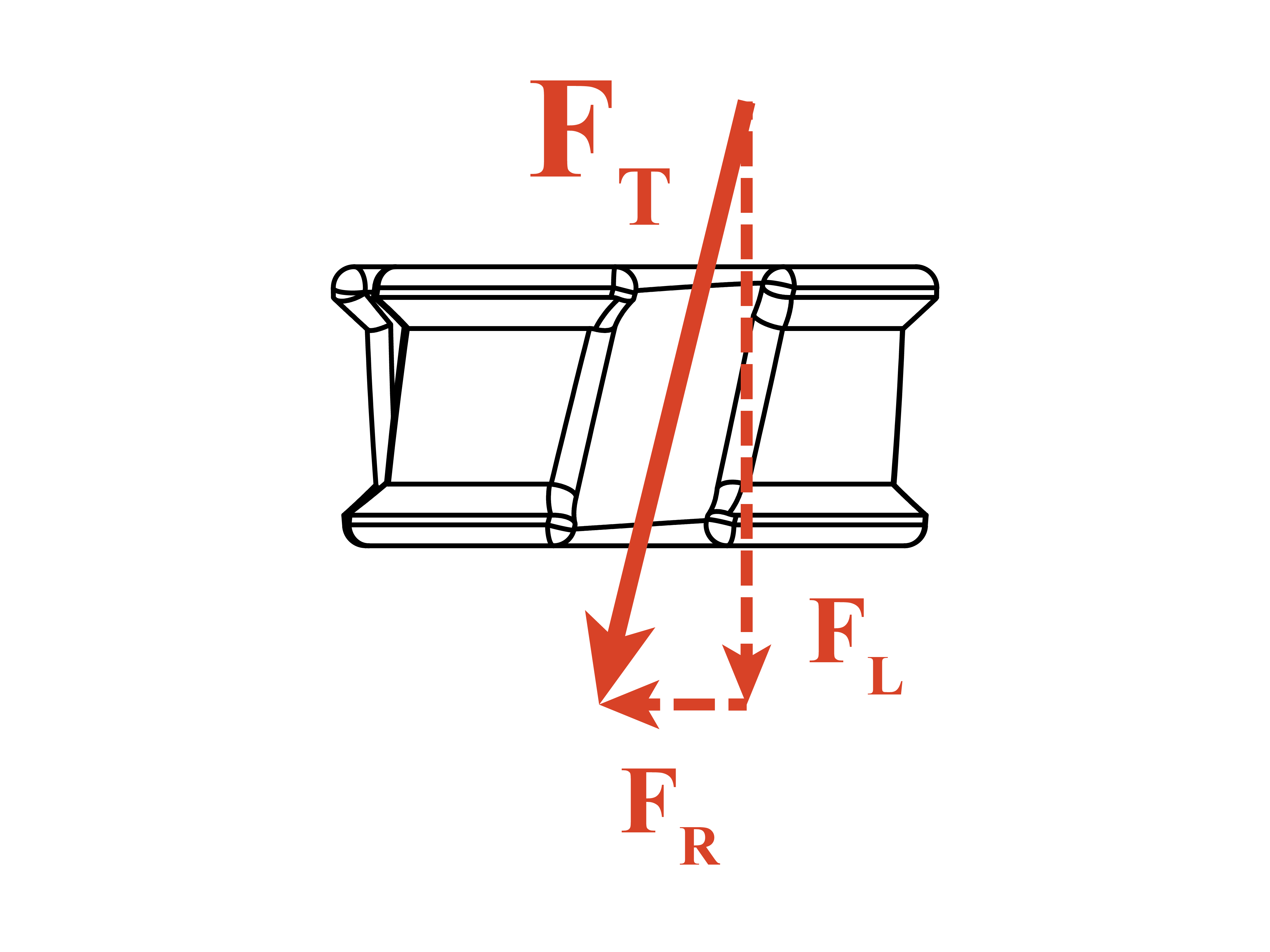}
        \caption{}
        \label{fig:fbd}
    \end{subfigure}
    \caption{(a) A closeup of the \robot{}'s McKibben actuators attached to the channels of the separator disks. (b) The separator disk for the \robot{} which is used to hold the four McKibben actuators in the desired helical orientation. (c) A diagram showing the tangential force, $F_T$ imparted by the interaction between the McKibben actuator and the separator disk, composed of a longitudinal force, $F_L$, along the body and a radial force, $F_R$, perpendicular to the body.}
\end{figure*}

Achieving sidewinding with serial modules, as with rigid-link snake robots, requires high spatial resolution in order to propagate a wave smoothly down the body. Achieving this high spatial resolution with soft actuators is difficult, as pneumatic actuation lines are much bulkier than the electrical lines powering and controlling servomotors. The principle behind the design of the \robot{} is to arrange the geometry of a set of McKibben actuators, replacing the high resolution of serial actuators with geometric placement of parallel actuators. 

Our design places the McKibben actuators in a helix around a set of spacers, attached so that they cross each spacer at an angle. When all the actuators are in their resting positions, they form a helix around a common central axis, whose pitch (turns per distance along the backbone) is dictated by the angle between the actuators and the spacer planes, as illustrated in \cref{fig:hissbot}. Contracting any one actuator forces the backbone into a helix (of larger diameter than the actuator resting helices), as illustrated in \cref{fig:constraintModel}. The phase of the backbone helix depends upon which actuator is contracted. Therefore, by cyclically contracting the actuators we can obtain a twisting-helix motion. As discussed in Hatton and Choset \cite{keyframe-wave}, twisting a helix in contact with a ground surface produces a sidewinding motion of the type illustrated in \cref{fig:sidewinding}.

One way to understand this helical contraction is by comparison with the McKibben arms discussed in Olson et al. \cite{mckibben-arm}, from which we have adapted the design of our snake. In that paper, the actuators are attached transversely to the disks and run parallel to the backbone, which bends in a planar arc when an actuator is contracted. In our case, the actuator itself twists around the backbone. This contraction in a helical curve results in the entire body becoming a ``macro-helix''. This twist is supported by the angled channels in the separator disks, producing a tangential component in the force the actuator supplies to the disk. This tangential force is illustrated in \cref{fig:fbd}.

Another means of viewing this geometry is to conceptualize the segments between each separator disk as modules in a serial-actuator snake robot. The identification of McKibben actuators across each body segment is a mechanical encoding of the phase relationships between rigid-link modules discussed in Tesch et al. \cite{tesch_parameterized_2009}. To better understand the setup, if we utilized straight, parallel actuators through each segment, and offset the axial phase at each disk, the snake would form multiple elbow-like segments which discretely approximate a helix. With the \robot{}, the twist throughout the body results in a continuous curvature change, forming the helical backbone we desire.

A third perspective on the geometry of the HISSbot is that when one of the helical actuators contracts, the backbone curls into a macro-helix geometry with the actuators lying on a helical shell around the backbone. As illustrated in \cref{fig:constraintModel}, this shape allows the shortened actuator to follow an ``inside curve” on the shell while the  other actuators follow ``outside curves”, preserving the compatibility constraint on the actuators at the disks.

\section{FABRICATION AND SETUP}

\begin{figure}[b]
    \centering
    \includegraphics[width=\columnwidth]{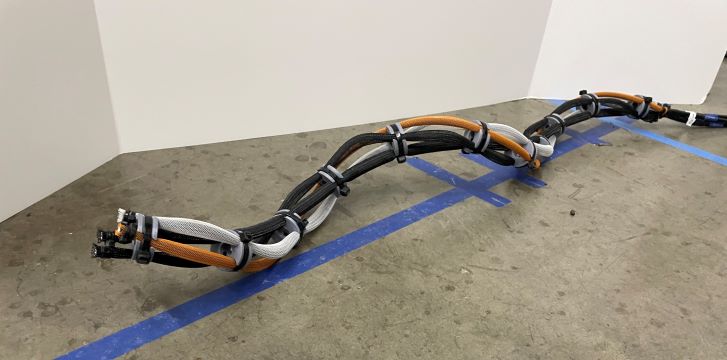}
    \caption{The physical model of the \robot{} with one actuator pressurized, showing a similar helical structure to the constraint model seen in \cref{fig:constraintModel}(b).}
    \label{fig:hissbot_helix}
\end{figure}

\begin{figure}
    \centering
    \includegraphics[width=\columnwidth]{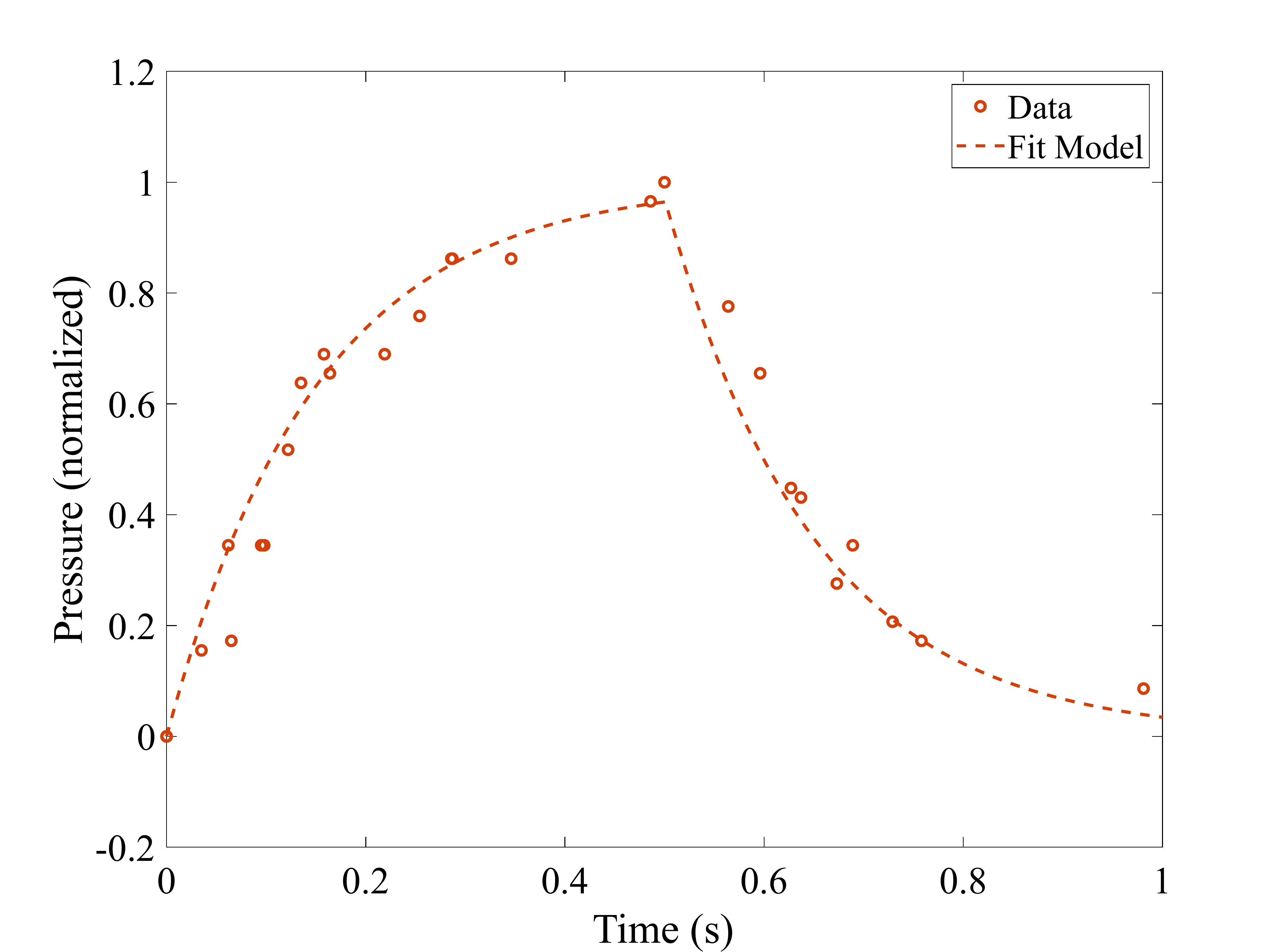}
    \caption{Experimental normalized pressure values from inflating and deflating the \robot{}, compared to the first-order system ($T=0.15$) fit to the data.}
    \label{fig:pressurization}
\end{figure}

Our physical instantiation of the \robot{} is composed of four McKibben actuators forming a quadruple helix, held in place by separator disks with angled channels. Each actuator is made from polyester sleeving (McMaster-Carr, 9284K412) and latex tubing (McMaster-Carr, 5234K972). The length of the \robot{} is 91cm. The outer diameter of the \robot{} is 3.2cm. The dimensions of the \robot{} is similar to other servo-chain snake robots. The actuators are then secured to separator disks with cable ties. The cable ties fully constrains the actuators to the separator disks, restricting any slip. The separator disks, as shown in \cref{fig:snakes,fig:braces}, are 3D printed using polylactic acid (PLA) thermoplastic to separate the actuators from one another. Each actuator is attached to a channel, angled at 20\degree{} between the holes at opposite faces, within each separator disk. This angled channel enforces the helical, geodesic structure on the cylindrical body. Each actuator of the \robot{} is connected to a manifold through air supply lines for pressurization.

To operate the \robot{}, we supply pressurized air to a manifold that distributes the air to eight solenoid valves. For each actuator, one solenoid acts as an inlet valve, and another as an outlet valve. These solenoids are connected to a relay module attached to a Raspberry Pi 4 Model B microcomputer. Actuation control is performed with Python on the Raspberry Pi.
 
Contraction of each actuator is linear as a function of pressure. Our initial tests confirmed behavior comparable to the standard McKibben model, as studied by Tondu and Lopez \cite{tondu_modeling_2000}. We determined that a pressure of 400kPa can be reliably achieved with minimal air leakage, and results in almost a 30\% contraction of the actuator. As such, 400kPa was used for all tests to ensure consistency without risk of actuator failure. By contracting any single actuator, we can induce a helical body shape in the \robot{}, as seen in \cref{fig:hissbot_helix}. This shape validates the shape demonstrated by the analytical model, shown in \cref{fig:constraintModel}. Additional work regarding the analytical model for the \robot{} is shown in \cite{fan_linear_2022}.


To support our design of control inputs for the \robot{} (discussed in the next section), we characterized the inflation time of the McKibben actuators. The inflation and deflation of an actuator follows the behavior of a first-order system with a time constant, $T=0.15$, as illustrated in in \cref{fig:pressurization}. Under square-wave pressure inputs, this inflation behavior results in shark-fin-like pressurization curves. We write the time period for an actuator to reach maximum pressure from atmospheric pressure as the inflation time, $\tau_I$. Using this value, we can parameterize gaits.


\section{SIDEWINDING IMPLEMENTATION}

With an understanding of the shape modes of the \robot{}, we then turn our attention to implementing sidewinding on the physical model.

\subsection{Actuation Methods}

\begin{figure}
    \centering
    \includegraphics[width=0.95\columnwidth]{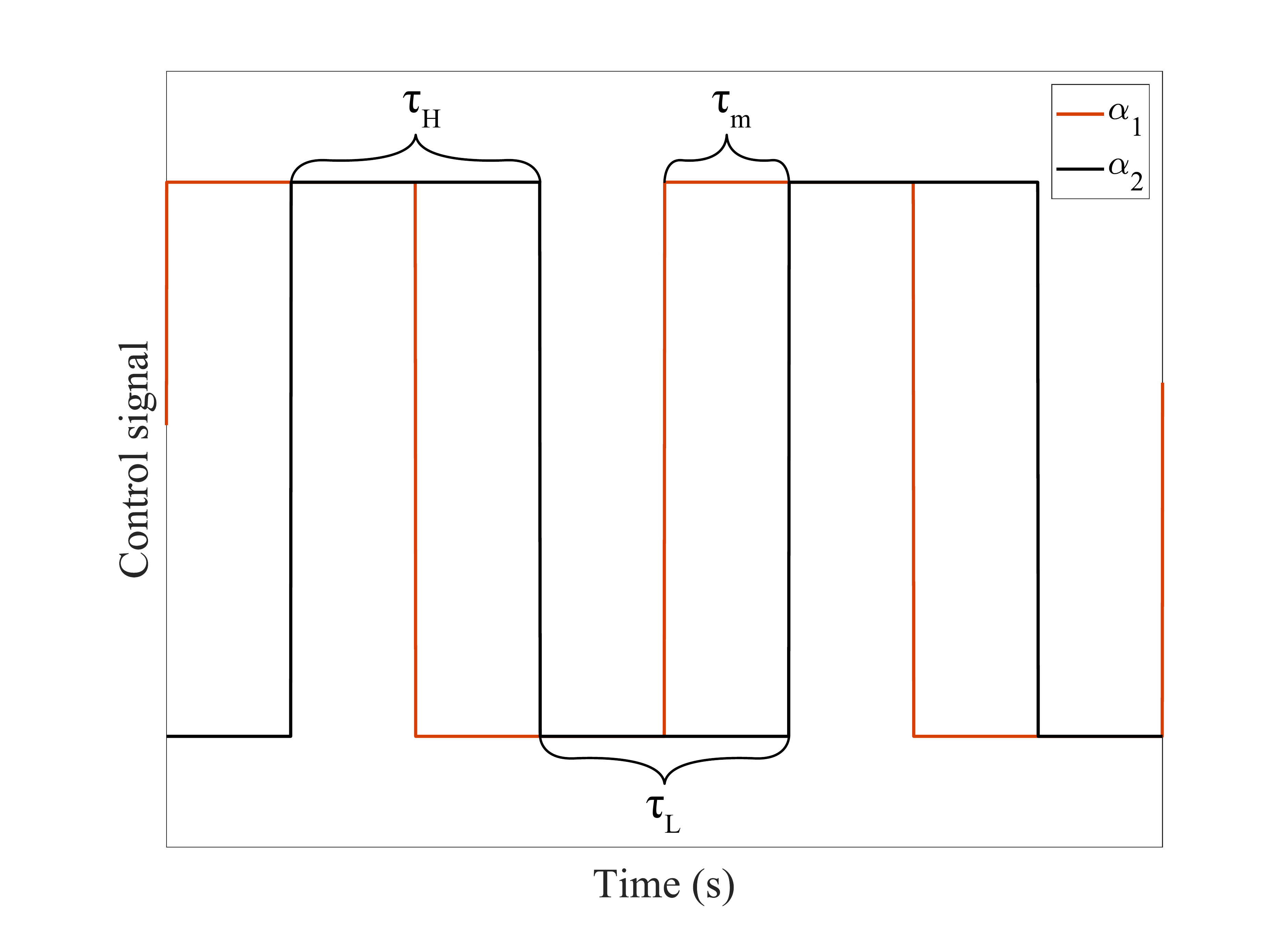}
        \caption{An example set of control signals for two actuators. $\tau_H$ denotes the length of time at the ``on'' position, $\tau_L$ denotes the length of time at the ``off'' position, and $\tau_m$ denotes the phase offset between the current and subsequent actuator.}
    \label{fig:controlSignal}
\end{figure}

\begin{figure*}
    \centering
        \includegraphics[width=2\columnwidth]{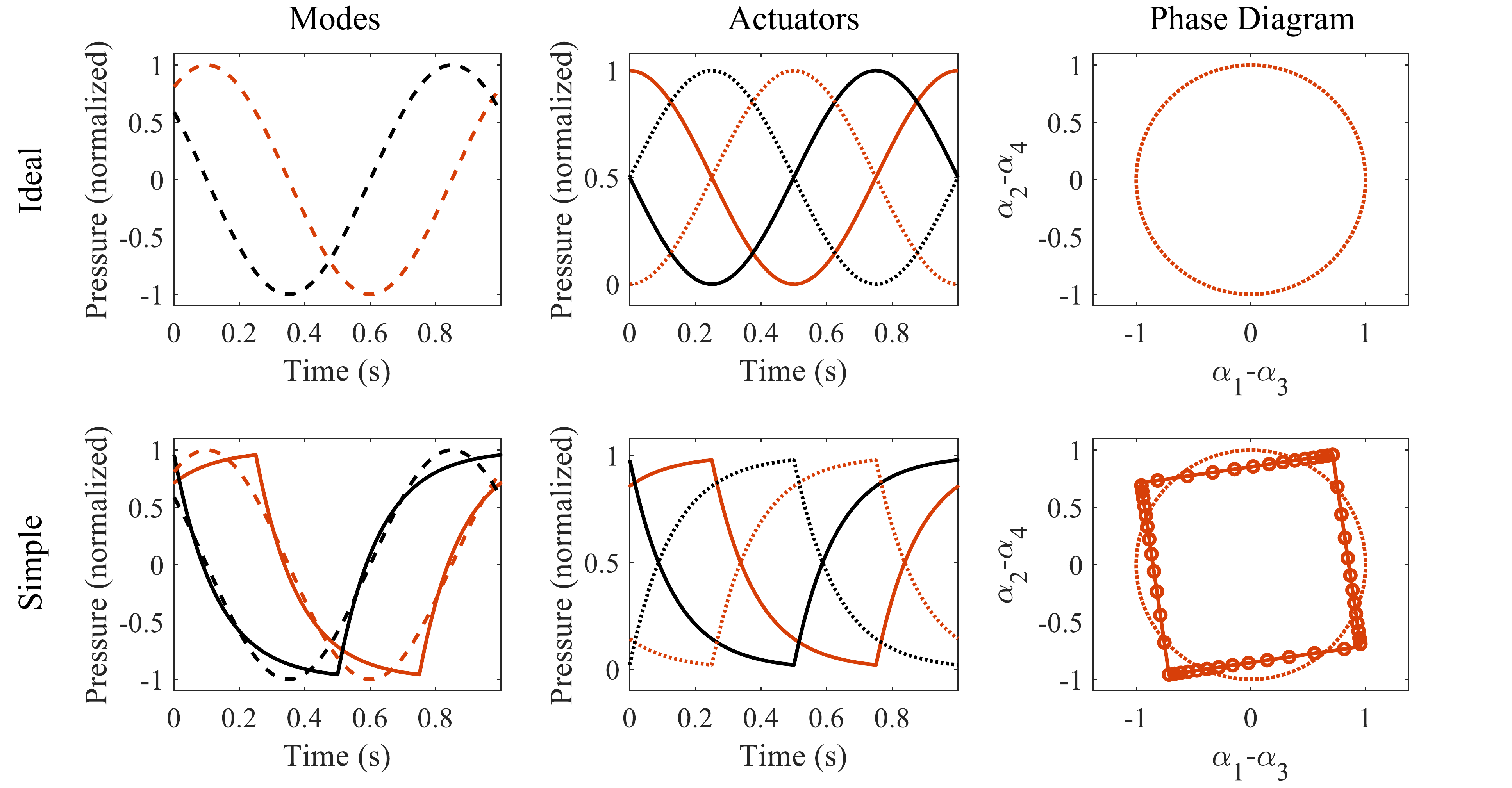}
    \caption{The top row represents the pressurization values using ideal sidewinding control. The bottom row represents the shark-fin-like pressurization values from the simple square-wave control, compared to the ideal values. The left column represents the pressurization cycle of each actuator pair. The middle column separates the actuator pairs to view the pressurization at the single actuator level. The middle column excludes the direct comparison between the ideal and simple values to better visualize the behavior. The right column presents a phase diagram view of this control, illustrating that the square wave control is equivalent of a square approximation of the circle representing sidewinding control.}
    \label{fig:actuation}
\end{figure*}

In servo-based snake robots, sidewinding is performed by feeding sinusoidal waves of actuation along the joints oriented in the dorsal and ventral directions, with a temporal offset of $\frac{\pi}{2}$ between the lateral and dorsal waves. The helical design of the \robot{} allows us to instantiate these traveling waves by actuating the McKibben actuators in sequence. For the input controls, we utilize a square wave determined by the length of time at the \emph{high} position, $\tau_H$, and the length of time at the \emph{low} position, $\tau_L$. To control the interaction between actuators, we then set an \emph{interactuator} delay between pairs of adjacent actuators, $\tau_m$. These values are illustrated in \cref{fig:controlSignal}. Sidewinding for the \robot{} is characterized by the following square-wave control:
\begin{equation}
    \begin{aligned}
        \tau_H &= \tau_L = \tau_I, \\
        \tau_m &= \frac{1}{2}\tau_I.
    \end{aligned}
\end{equation}
We define two actuator pairs, $(\alpha_1,\alpha_3)$ and $(\alpha_2,\alpha_4)$, composed of two helical McKibben actuators on opposing sides of a separator disk. In order to induce a rolling helical motion with the \robot{}, we start with sinusoidal waves for each actuator pair, temporally offset by $t_o = \frac{\pi}{2}$:
\begin{equation}
    \begin{aligned}
        \alpha_1-\alpha_3 &= \cos(t), \\
        \alpha_2-\alpha_4 &= \cos(t-t_o).
    \end{aligned}
\end{equation}
The top left plot in \cref{fig:actuation} shows an example of the pressurization cycles of the actuator pairs used to induce this sidewinding motion. At the individual actuator level, this sidewinding motion is equivalent to sinusoidal waves which range in amplitude from 0 to 1. Implementing a simple control scheme of using square waves to control each actuator results in a shark-fin-like pressurization profile.

Looking at the actuator pair profile, we obtain a shark-fin wave ranging from -1 to 1, which closely resembles the ideal sinusoidal actuation. The bottom left plot in \cref{fig:actuation} shows the pressurization result of this simple square-wave control. The sinusoidal actuation is further illustrated by viewing the phase diagram of the actuator pairs, illustrated in the right column of \cref{fig:actuation}. The square-wave control demonstrates a square approximation of the circle, representing the sidewinding behavior.

The supplementary video shows an example of how the \robot{} sidewinds in various media. The performance of the \robot{} is determined by its displacement from sidewinding. In order to normalize the results, the overall displacement is measured at the center of mass (COM) of the body. The displacement is then normalized by the body length and number of cycles in the test to obtain the desired metric of body lengths per cycle (BL/cycle). These units will assist with comparison among future iterations of the \robot{}. This metric does differ from the typical meters per second (m/s) used in literature. BL/cycle represents a more agnostic measurement of performance, which can be used more easily to compare among other snake-like robots of varied sizes.

\begin{figure}
    \centering
        \includegraphics[width=1\columnwidth]{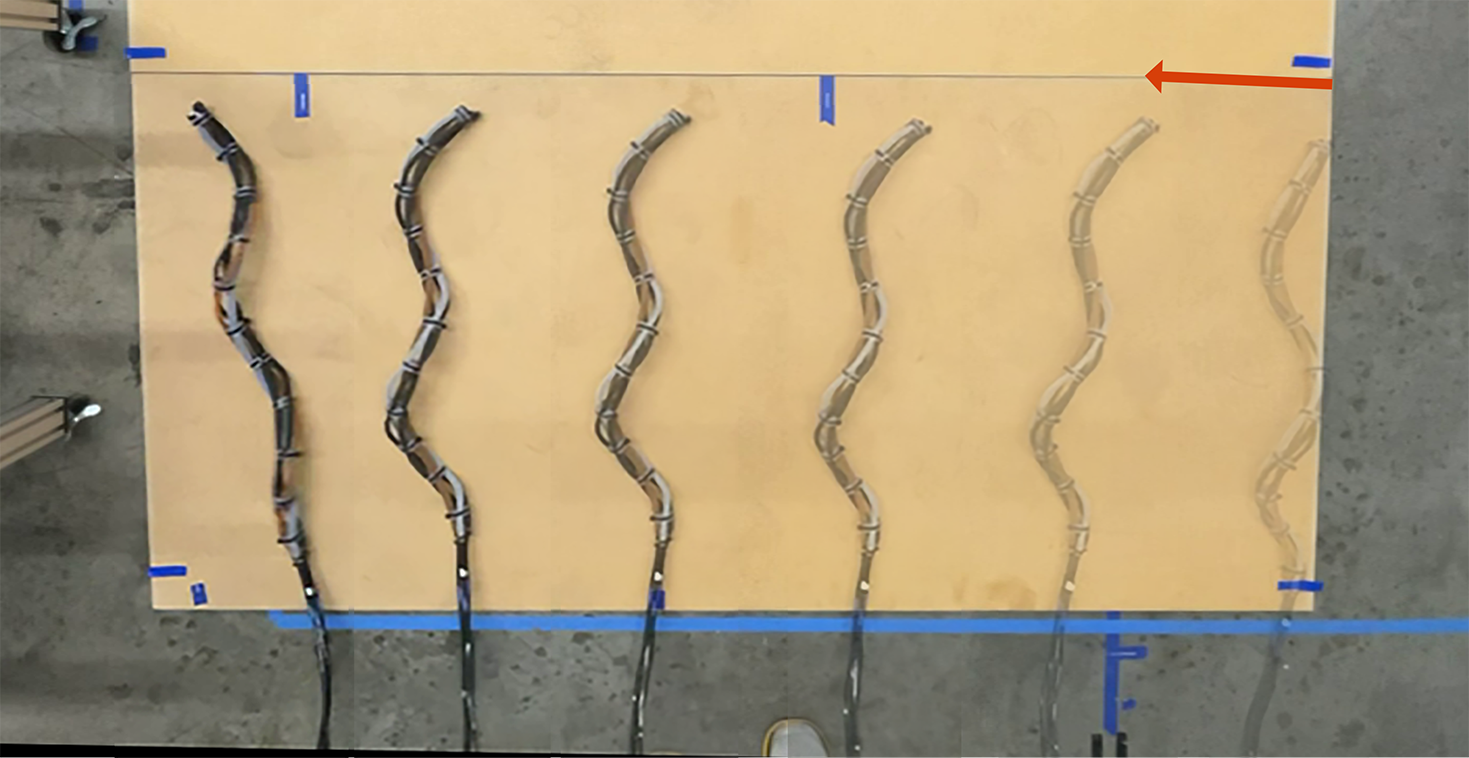}
    \caption{An example of the \robot{} movement across a wooden surface over the span of a single experiment. The orange arrow shows the movement of the body over time. Each body is separated by 2 cycles of the sidewinding gait. The supplementary video includes sidewinding across the different media tested.}
    \label{fig:sidewindingtime}
    \vspace{-10pt}
\end{figure}

Results from \sidewind{} validated that the actuation method creates a sidewinding motion. \cref{fig:sidewindingtime} shows the movement of the \robot{} across a wooden surface, exhibiting similar sidewinding behavior demonstrated by \cref{fig:sidewinding}.

\subsection{Surface Tests}

One of the benefits of sidewinding is its insensitivity to surface materials. The minimal slip seen with a looping body and static ground contacts also makes it suitable for traversing loose terrain. Therefore, we tested the sidewinding action of the \robot{} on low-friction and loose material. We cycle through the motion ten times to capture any long-term error based on the number of cycles. The surfaces tested are listed in \cref{tab:surfaceresults}. For the granular media, only six-cycle tests were conducted due to limited area of the test bed.

We additionally performed tests of the sidewinding control in the opposite order. By doing so, we demonstrate that the \robot{} sidewinds in the opposing direction as expected. \cref{tab:surfaceresults} shows no significant differences in average displacement across solid surfaces. A significant reduction in displacement is seen with the tests on granular media. The granular media acts as a lower friction surface, and the rapid changes in pressure results in larger slip compared to the solid surfaces.

\begin{table}[]
    \caption{Average displacement of the \robot{} after sidewinding tests with varied surfaces.}
    \label{tab:surfaceresults}
    \centering
    \begin{tabular}{c c c}
        \multicolumn{1}{c}{\begin{tabular}[c]{@{}c@{}}Surface medium\end{tabular}} & \multicolumn{1}{c}{\begin{tabular}[c]{@{}c@{}}Sidewinding\\displacement\\ (BL/cycle)\end{tabular}} & \multicolumn{1}{c}{\begin{tabular}[c]{@{}c@{}}Reversed sidewinding\\dispacement\\ (BL/cycle)\end{tabular}} \\ \hline\hline
        Polished Concrete & $0.19\pm0.02$ & $0.17\pm0.01$ \\
        Wood & $0.18\pm0.01$ & $0.16\pm0.01$ \\ 
        Rubber & $0.19\pm0.01$ & $0.17\pm0.01$ \\
        Granular Media & $0.14\pm0.01$ & $0.13\pm0.01$ \\
        \hline\hline
    \end{tabular}%
\end{table}

\subsection{Constrained Environments}

On an open, unconstrained ground, executing sidewinding control on the \robot{} results in expected sidewinding behavior. In constrained environments, the same form of control results in different, yet valuable, forms of locomotion. As shown in prior work done by Tesch et al. \cite{tesch_parameterized_2009} and Hatton and Choset \cite{keyframe-wave}, the same control that results in sidewinding can also be used in more constrained environments, pipes and narrow passages. We tested sidewinding control on the \robot{} within a pipe, a narrow passage, and along a wall. \cref{tab:constrainedresults} shows the displacement through these environments. The supplementary video includes the resulting locomotion through the varying constrained environments. Notably, although the average displacement per cycle decreased compared to sidewinding, the \robot{} was still able to utilize the same control to navigate the terrain.

\begin{table}[]
    \caption{Average displacement of the \robot{} after sidewinding control test with varying constrained environments.}
    \label{tab:constrainedresults}
    \centering
    \begin{tabular}{c c}
        \multicolumn{1}{c}{\begin{tabular}[c]{@{}c@{}}Constrained environment\end{tabular}} & \multicolumn{1}{c}{\begin{tabular}[c]{@{}c@{}}Sidewinding displacement\\ (BL/cycle)\end{tabular}} \\ \hline\hline
        Within a pipe (6.5cm diameter) & $0.02\pm0.01$\\
        Narrow Passage (6.5cm wide) & $0.02\pm0.01$\\ 
        Along a wall & $0.03\pm0.01$\\
        \hline\hline
    \end{tabular}%
\end{table}


\section{DISCUSSION AND FUTURE WORK}

Sidewinding is a key gait for snake robots. As discussed in Tesch et al. \cite{tesch_parameterized_2009}, sidewinding is the second fastest of the standard snake robot gaits. Additionally, it is more robust to terrain variations than rolling (the fastest snake robot gait), while also providing a more stable platform for sensors---The twisted-helix motion keeps the general orientation of the body more consistent to the ground, compared to the constant rolling body of the rolling gait. To date, sidewinding has only been demonstrated on snake robots with high-resolution serial actuation.

In this paper, we presented a parallel-helix actuator design that achieves sidewinding with two active control modes. We instantiated the design as the \robot{}. The instantiation of the design demonstrates the feasibility of having a parallel-helix design for sidewinding.

Building on this proof-of-concept, our future work with the \robot{} will include improvements to the physical platform, along with more fundamental exploration of concepts in locomotion. Early investigation of the \robot{} architecture suggests avenues for physical design improvements. Integrating onboard valves on the robot can reduce the number of air tethers required for operation. Given the current configuration of the \robot{}, a key limit of its mobility is the drag on the tether. Mitigating this factor would provide greater control of the robot itself. We currently bypass this issue by lifting the tether during the tests to focus on the motion of the mechanism itself. In addition, replacing the current solenoid valves with pressure regulators would create a smoother execution of the desired sidewinding gait, and minimize the slip on a given surface. These design improvements would make the \robot{} a more suitable platform for integration of sensor suites for various applications. We have performed an initial integration test with camera and radiation spectrometer modules for inspection of radioactive spaces, discussed in detail in \cite{wilson_modular_2022}.

Thus far, we have only explored the actuation pattern that produces sidewinding movement. Exploring additional capabilities of the robot is vital to creating a viable alternative to the servo-based snake robots. We are interested in developing other patterns of actuation with the current geometry to induce longitudinal movement. A possible pattern is a synchronized cyclical actuation of the actuator pairs to form a gait resembling lateral undulation. Additionally, investigating other modal geometries can assist with traversing more complicated terrain. For instance, designing elliptical separator disks to induce a flattened helix. Prior research has shown that flattening the helical structure allows a snake robot to effectively sidewind through granular slopes \cite{sidewinding-on-slopes, marvi_sidewinding_2014}.

At a broader level, the HISSbot design illustrates the role of shape modes in locomotion. Moving forward, we aim to use the HISSbot as a platform for further investigating modal choice, and to study the use of ``steering actuators" to redirect the action of more powerful actuators. The study in Ma et al. \cite{ma_controlled_2013}, presents an insect-sized flapping wing robot, with separate controls for powering the flapping motion and guiding the angle of the wing. Inspired by this study, we are exploring different types of actuation to add to the \robot{} for improved control. We are in the process of developing separator disks with adjustable channel angles. The extra control gained from adjusting the angle would allow us to switch between helical and flat geometries, to perform sidewinding, rolling, or lateral undulation depending on the terrain. 


\section{ACKNOWLEDGEMENTS}

This work was funded in part by DOE/NNSA Office of Defense Nuclear Nonproliferation Research and Development and by the National Science Foundation under awards 1653220 and 1826446.


\bibliographystyle{ieeetr}
\bibliography{ref}

\end{document}